\documentclass[11pt]{article}

\usepackage[preprint]{acl}

\usepackage{times}
\usepackage{latexsym}

\usepackage[T1]{fontenc}

\usepackage[utf8]{inputenc}

\usepackage{microtype}

\usepackage{inconsolata}

\usepackage{times}
\usepackage{latexsym}
\usepackage{booktabs}
\usepackage{multirow}
\usepackage{array}
\usepackage{xcolor}
\usepackage[T1]{fontenc}
\usepackage[utf8]{inputenc}
\usepackage{microtype}
\usepackage{inconsolata}
\usepackage{graphicx}
\usepackage{amsmath}
\usepackage{amssymb}
\title{GoiBERT: Compact Vietnamese Language Models via Two-Stage Knowledge Distillation}

\author{
  \textbf{Quang Anh Kiet Tran\textsuperscript{1,3}\thanks{\ \ Equal contribution.}} \quad
  \textbf{Nam Anh Tran\textsuperscript{2,3}\footnotemark[1]} \quad
  \textbf{Manh Hung Trinh\textsuperscript{4}} \\
  \textbf{Thanh Danh Le\textsuperscript{2,3}} \quad
  \textbf{Khang Nhat Hoang Vo\textsuperscript{5}} \quad
  \textbf{Huynh Dai Nhan Tran\textsuperscript{6}} \\[0.8em]
  \textsuperscript{1}UIT, VNU-HCM \quad
  \textsuperscript{2}IU, VNU-HCM \quad
  \textsuperscript{3}TwoHearts Technologies Co., Ltd., Ho Chi Minh City, Vietnam \\
  \textsuperscript{4}University of Eastern Finland \quad
  \textsuperscript{5}MBZUAI \quad
  \textsuperscript{6}University of Jyv\"askyl\"a \\[0.3em]
  {\small \textbf{Correspondence:} \href{mailto:huynh.d.n.tran@jyu.fi}{huynh.d.n.tran@jyu.fi}}
}

\begin{document}
\maketitle
\begin{abstract}
Vietnamese pre-trained encoders have improved rapidly, but the
strongest ones are large, multilingual-vocabulary models or depend on
external word segmentation, which makes them costly to deploy.
We present \textbf{GoiBERT}, a family of two compact Vietnamese
encoders with 26M and 49M parameters that operate directly on raw text.
GoiBERT learns from two complementary teachers.
A same-tokenizer Vietnamese encoder first teaches general-purpose
representations.
A much larger teacher then provides task-level guidance through soft
labels that are generated without ever labeling its own training data.
Across eight benchmarks covering text classification, aspect-based
sentiment analysis, natural language inference, and named entity
recognition, GoiBERT-large achieves the highest average score among
non-teacher models.
It does so with fewer than half the parameters of the strongest compact
baseline and twice its throughput.
GoiBERT-base retains 99\% of that baseline's performance at a quarter
of its size.
Our analysis shows that compression is nearly lossless for sentiment
and classification tasks but remains costly for inference and entity
recognition, which points to clear directions for future work. Our code and models are available at: \url{https://github.com/Nhantran65/GoiBERT.git}
\end{abstract}

\section{Introduction}
\label{sec:intro}

Pre-trained Transformer encoders remain the workhorse of practical language
understanding: they power text classification, sentiment analysis,
information extraction, and retrieval in production systems where
generative large language models are too slow or too costly to serve
\cite{devlin2019bert,liu2019roberta,conneau2020xlmr,warner2025smarter}.
Vietnamese NLP has followed this trajectory closely, moving from
multilingual encoders to a growing family of monolingual models such as
PhoBERT, ViDeBERTa, ViSoBERT, and the recent BamiBERT, as well as
CafeBERT, a large XLM-R model further adapted to Vietnamese
\cite{nguyen2020phobert,tran2023videberta,nguyen2023visobert,nguyen2026bamibert,do2024vlue}.

These models have steadily raised the bar on Vietnamese benchmarks, but
each comes with a deployment cost.
The strongest model, CafeBERT, has 559M parameters, and more than
45\% of them sit in its multilingual embedding matrix.
PhoBERT expects word-segmented input and therefore needs an external
segmenter at inference time \cite{vu2018vncorenlp}.
ViSoBERT targets social-media text, and even the most compact strong
monolingual baseline, BamiBERT, uses a 102M-parameter encoder.
For the many Vietnamese applications that run on CPUs, small GPUs, or
under tight latency budgets, the field still lacks a \emph{compact},
\emph{segmentation-free} encoder that holds up across diverse tasks.

\begin{figure}[t]
  \centering
  \includegraphics[width=\columnwidth]{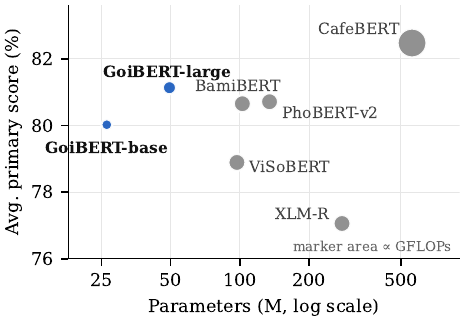}
  \caption{Average score over eight Vietnamese benchmarks versus model
  size. Both GoiBERT variants lie on the accuracy--efficiency frontier
  of non-teacher models: GoiBERT-large has the highest average with
  less than half of BamiBERT's parameters, and GoiBERT-base stays close
  with about a quarter of them.}
  \label{fig:teaser}
\end{figure}

Knowledge distillation (KD) is the standard tool for closing this gap in
English, where compact students such as DistilBERT, TinyBERT, MobileBERT,
and MiniLM retain most of their teachers' accuracy at a fraction of their
cost \cite{hinton2015distilling,sanh2019distilbert,jiao2020tinybert,sun2020mobilebert,wang2020minilm}.
Transferring these recipes to Vietnamese, however, raises a design
question that English work rarely faces: \emph{which teacher should the
student learn from?}
The most accurate Vietnamese teacher (CafeBERT) uses a large
multilingual vocabulary that does not match any compact Vietnamese
tokenizer.
This rules out the token-level hidden-state and attention matching
that makes TinyBERT-style general distillation effective.
A same-tokenizer teacher (BamiBERT) supports such matching but is
itself weaker on several tasks.
To our knowledge, no prior work has studied how to combine these
complementary signals for Vietnamese, and Vietnamese compact encoders
have not been evaluated systematically for both accuracy and deployment
cost.

We present \textbf{GoiBERT}, a family of two compact Vietnamese encoders:
GoiBERT-base (26.4M parameters) and GoiBERT-large (49.4M parameters).
Our key idea is to \emph{assign each teacher to the stage where its
strengths matter}.
In the first stage, BamiBERT, which shares GoiBERT's tokenizer, teaches
general-purpose representations through layer-wise hidden-state and
attention transfer on unlabeled text.
In the second stage, CafeBERT supplies task-specific soft labels at the
output level, where tokenizer mismatch is irrelevant.
We produce these labels \emph{out of fold}, so that no teacher ever
labels an example it was trained on.
As Figure~\ref{fig:teaser} shows, the resulting students move the
accuracy--efficiency frontier for Vietnamese.
GoiBERT-large achieves the best average score among non-teacher models
with fewer than half of BamiBERT's parameters and twice its GPU
throughput.
GoiBERT-base retains 99\% of BamiBERT's average score with a quarter of
its parameters and 3.5$\times$ its throughput.

Our contributions are threefold:
\begin{itemize}
  \setlength\itemsep{1pt}
  \item \textbf{Models.} We release GoiBERT-base and GoiBERT-large,
  compact Vietnamese encoders that process raw text without word
  segmentation and offer two complementary accuracy--efficiency
  operating points.
  \item \textbf{Method.} We propose a two-teacher, two-stage
  distillation framework. It pairs representation transfer from a
  tokenizer-compatible teacher with out-of-fold, output-level
  supervision from a stronger cross-tokenizer teacher, and it combines
  this with teacher-derived embedding initialization and layer-wise
  learning-rate decay.
  \item \textbf{Analysis.} We evaluate on eight benchmarks (15 metrics,
  three seeds) and measure parameters, FLOPs, throughput, memory, and
  CPU latency. Ablations and per-task analysis show where compression
  is nearly free (sentiment, ABSA, topic, and spam classification) and
  where it still costs accuracy (NLI and NER).
\end{itemize}

\section{Related Work}
\label{sec:related}

Our work draws on three lines of research: Vietnamese pre-trained
encoders, knowledge distillation for compact Transformers, and
efficient language-specific models.

\paragraph{Vietnamese pre-trained encoders.}
Vietnamese NLP has moved from multilingual encoders such as XLM-R
\cite{conneau2020xlmr} to a growing family of monolingual models.
After early BERT and ELECTRA adaptations \cite{bui2020improving},
PhoBERT set strong baselines but requires word-segmented input from an
external segmenter \cite{nguyen2020phobert,vu2018vncorenlp}.
This design choice reflects a property of the language: Vietnamese is
written as space-separated syllables, and many words span several of
them, so a model must either receive word boundaries explicitly or learn
them from raw text.
Later models changed the pre-training objective (ViDeBERTa;
\citealp{tran2023videberta}) or the target domain (ViSoBERT for social
media and ViHealthBERT for health;
\citealp{nguyen2023visobert,minh2022vihealthbert}).
Others added linguistic inductive bias: HuTieuBERT injects
morpheme-boundary information into a pre-trained encoder to reflect the
isolating structure of Vietnamese \cite{dinh2026morphology}.
BamiBERT is a recent general-purpose raw-text encoder
\cite{nguyen2026bamibert}, while CafeBERT continues pre-training
XLM-R-large on Vietnamese and leads the VLUE benchmark
\cite{do2024vlue}.
Generative models such as ViT5, BARTpho, and PhoGPT
\cite{phan2022vit5,tran2022bartpho,nguyen2023phogpt} target text
generation and are complementary to encoders, which remain the
cost-effective choice for understanding tasks.
Benchmarks such as VLUE and ViGLUE \cite{do2024vlue,tran2024viglue}
have made these models comparable, but only in terms of accuracy.
Similar efforts for other medium-resource languages, such as
Indonesian, Indic languages, Bangla, and Thai, have likewise focused on
full-size models
\cite{wilie2020indonlu,kakwani2020indicnlpsuite,bhattacharjee2022banglabert,lowphansirikul2021wangchanberta}.
This line of work has therefore steadily raised accuracy while model
sizes have stayed around or above 100M parameters.
GoiBERT asks the complementary question of how much of that accuracy
can be retained at a fraction of the cost.

\paragraph{Knowledge distillation for compact encoders.}
Knowledge distillation (KD) trains a small student to mimic the
softened predictions of a larger teacher \cite{hinton2015distilling}.
For BERT-style encoders, KD has been extended from output logits to
intermediate signals.
PKD and DistilBERT match intermediate layers or initialize the student
from the teacher \cite{sun2019patient,sanh2019distilbert}, MobileBERT
transfers layer-wise features \cite{sun2020mobilebert}, and MiniLM
transfers self-attention distributions
\cite{wang2020minilm,wang2021minilmv2}.
TinyBERT \cite{jiao2020tinybert} combines these signals into the
two-stage scheme we build on: general distillation of hidden states and
attention on unlabeled text, followed by task-specific distillation.
The general stage lets a compact student acquire broad language
knowledge without being pre-trained from scratch, and the task stage
then specializes it with the teacher's predictions.
Comparative studies further show that pre-training the student
\cite{turc2019wellread} and the choice of distillation objective
\cite{udagawa2023comparative} strongly affect the final model.
Two findings from this literature shape our design.
First, a teacher much larger than its student can transfer poorly
\cite{cho2019efficacy,mirzadeh2020ta}, which favors a mid-sized teacher
for representation transfer.
Second, soft labels partly act as a regularizer, so a student can match
or even surpass its teacher \cite{furlanello2018born,yuan2020revisiting}.
Existing two-stage methods use a single teacher in both stages, and
multi-teacher approaches combine several teachers into one supervision
signal \cite{liu2019mtdnnkd,yuan2021reinforced,avram2022romanian}.
When teacher and student vocabularies differ, distillation must
additionally align them \cite{boizard2025uld,zhang2024dskd}.
GoiBERT sidesteps this problem by giving each teacher its own stage.
A same-tokenizer teacher supervises representations, and a
cross-tokenizer teacher supervises only task outputs, whose soft labels
we generate out of fold in the spirit of stacked generalization
\cite{wolpert1992stacked}.

\paragraph{Efficient language-specific encoders.}
A complementary line of work builds compact models for a single
language.
One route starts from multilingual models, either trimming their
vocabulary \cite{abdaoui2020loadwhat,ushio2023vocabtrim} or distilling
them into language-specific students
\cite{mukherjee2020xtremedistil,ansell2023bistil,cruz2025extracting}.
Another distills monolingual teachers, as in DistilCamemBERT for French
\cite{delestre2022distilcamembert}.
In both cases, the student typically inherits its vocabulary and its
supervision from the same single teacher.
GoiBERT combines the two routes.
Its tokenizer and initial embeddings come from a monolingual teacher,
in line with evidence that initializing small models from larger ones
helps \cite{samragh2023subcloning,xu2024weightselection}.
Its task supervision comes from a teacher of multilingual origin.
Pruning \cite{sanh2020movement,xia2022cofi}, quantization
\cite{kim2021ibert}, and parameter sharing \cite{lan2020albert} are
orthogonal and can be applied on top of
GoiBERT.
Finally, parameter counts and FLOPs can misrepresent real speed
\cite{dehghani2022efficiency}.
Following recent efficiency-focused encoders such as ModernBERT
\cite{warner2025smarter}, we therefore report measured throughput,
memory, and latency alongside predictive accuracy.

\section{GoiBERT}
\label{sec:method}

\begin{figure*}[t]
  \centering
  \includegraphics[width=0.80\textwidth]
  {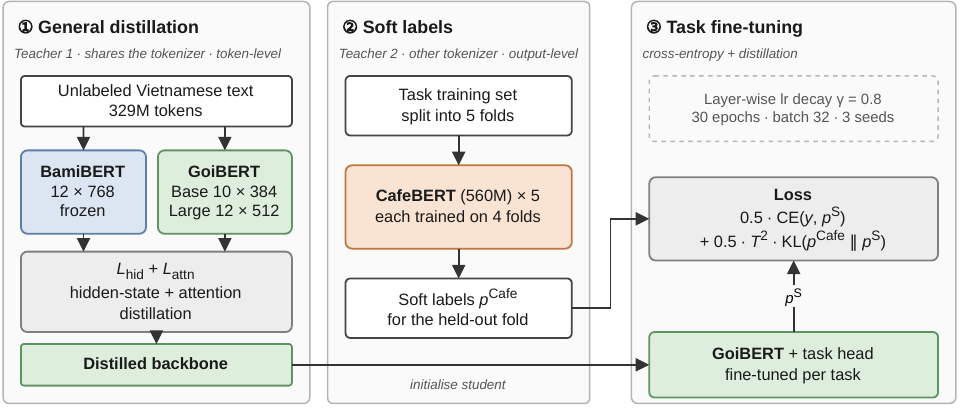}
  \caption{Overview of the GoiBERT training pipeline.
  \textbf{(1)}~General distillation: the student matches the hidden
  states and attention maps of a frozen BamiBERT, which shares its
  tokenizer, on unlabeled Vietnamese text.
  \textbf{(2)}~Out-of-fold soft labels: five CafeBERT teachers, each
  fine-tuned on four folds, label the held-out fold, so that no
  example is labeled by a teacher that saw it.
  \textbf{(3)}~Task fine-tuning: the distilled backbone is trained on
  gold labels and CafeBERT soft labels with layer-wise learning-rate
  decay.}
  \label{fig:training_pipeline}
\end{figure*}

GoiBERT is designed around one observation.
The two strongest available teachers offer \emph{different kinds} of
knowledge, and each kind can only be transferred at a particular
granularity.
BamiBERT tokenizes text exactly as GoiBERT does, so its token-level
hidden states and attention maps can be matched position by position.
CafeBERT is considerably more accurate on several tasks, but its
250K-entry multilingual vocabulary segments the same sentence
differently, so its knowledge can only be transferred cleanly through
sequence- or word-level predictions.
GoiBERT therefore uses BamiBERT for \emph{task-agnostic representation
transfer} and CafeBERT for \emph{task-specific output supervision}
(Figure~\ref{fig:training_pipeline}), extending the two-stage design of
TinyBERT \cite{jiao2020tinybert} to two heterogeneous teachers with complementary roles.

\subsection{Compact Architecture}
\label{sec:arch}

GoiBERT follows the RoBERTa encoder architecture \cite{liu2019roberta}
and reuses BamiBERT's 20,481-token vocabulary and tokenizer, so it
processes raw Vietnamese text without word segmentation.
\textbf{GoiBERT-base} has 10 Transformer layers, hidden size 384,
6 attention heads, and a 1,536-dimensional feed-forward layer
(26.4M parameters).
\textbf{GoiBERT-large} has 12 layers, hidden size 512, 8 heads, and a
2,048-dimensional feed-forward layer (49.4M parameters).
Both use 64-dimensional attention heads, as BamiBERT does.
The compact vocabulary is itself a source of efficiency.
Embeddings account for only 33\% and 23\% of the parameters of
GoiBERT-base and GoiBERT-large, respectively, compared with 46\% for
CafeBERT and 69\% for XLM-R (Table~\ref{tab:full}).
To inherit lexical knowledge from the teacher, we initialize
GoiBERT-large's word embeddings by projecting BamiBERT's 768-dimensional
embedding matrix onto its 512 leading principal components and
rescaling to the student's initialization scale.
Transformer blocks are initialized randomly.
Appendix~\ref{app:architecture} gives a layer-level diagram.

\subsection{Stage 1: General Distillation}
\label{sec:gd}

In the first stage, the student learns from a frozen BamiBERT on
unlabeled text by matching its intermediate representations.
Let $v\in\{\mathrm{base},\mathrm{large}\}$ denote a variant with $L_v$
layers and hidden size $d_v$.
Let $H^v_\ell\in\mathbb{R}^{n\times d_v}$ be the student's hidden states
at layer $\ell$ ($\ell=0$ is the embedding output), and let
$H^B_{\pi_v(\ell)}\in\mathbb{R}^{n\times 768}$ be those of the teacher
layer paired with it by the mapping $\pi_v$.
GoiBERT-large has 12 layers like BamiBERT, so $\pi_{\mathrm{large}}$ is
the identity.
GoiBERT-base has only 10 layers, which rules out a one-to-one pairing.
Instead, $\pi_{\mathrm{base}}$ is an order-preserving mapping that pairs
the embedding outputs and selects 10 of the teacher's 12 layers as
targets, so that lower student layers learn from lower teacher layers
and upper from upper \cite{jiao2020tinybert}
(Appendix~\ref{app:distillation}).

Because the hidden sizes differ, a trainable linear projection
$P^v_\ell$ maps student states into the teacher space:
\begin{equation}
  \mathcal{L}_{\mathrm{hid}}
  = \frac{1}{L_v+1}\sum_{\ell=0}^{L_v}
  \operatorname{MSE}_{\mathrm{nonpad}}\!\bigl(P^v_\ell(H^v_\ell),\,H^B_{\pi_v(\ell)}\bigr),
  \label{eq:hid}
\end{equation}
where the error is computed only over non-padding positions.

The two models also have different numbers of attention heads, so
head-to-head matching is undefined.
We instead match head-averaged attention distributions
$\bar{A}^m_\ell=\frac{1}{a_m}\sum_{h=1}^{a_m}A^m_{\ell,h}$, where
$A^m_{\ell,h}$ is the post-softmax attention matrix of head $h$ in
model $m$:
\begin{equation}
  \mathcal{L}_{\mathrm{att}}
  = \frac{1}{L_v}\sum_{\ell=1}^{L_v}
  \operatorname{MSE}\!\bigl(\bar{A}^v_\ell,\,\bar{A}^B_{\pi_v(\ell)}\bigr).
  \label{eq:att}
\end{equation}
Averaging keeps the objective architecture-agnostic while still
transferring \emph{where} each layer attends
\cite{jiao2020tinybert,wang2020minilm}.
The general-distillation objective is
\begin{equation}
  \mathcal{L}_{\mathrm{GD}} = \mathcal{L}_{\mathrm{hid}} + \mathcal{L}_{\mathrm{att}}.
  \label{eq:goibert-general-loss}
\end{equation}

\subsection{Stage 2: Out-of-Fold Task Distillation}
\label{sec:td}

In the second stage, the distilled backbone is fine-tuned on each
downstream task with gold labels and soft labels from CafeBERT.
A naive approach would fine-tune CafeBERT on the full training set and
then label that same set.
However, a 559M-parameter teacher tends to fit its training examples
closely, so its predictions on them become nearly one-hot and carry
little of the ``dark knowledge'' that makes soft labels useful
\cite{hinton2015distilling,stanton2021does}.
We therefore adopt a cross-fitting scheme inspired by stacked
generalization \cite{wolpert1992stacked}.
We split the training set into five folds, fine-tune one CafeBERT on
each set of four folds, and use it to predict only the held-out fold.
Every training example thus receives a soft target from a teacher that
has never seen it, which reflects the teacher's genuine uncertainty on
that example.

Given gold label $y$, student logits $z_S$, teacher logits $z_C$, and
$p^{(T)}_m=\operatorname{softmax}(z_m/T)$, the task objective is
\begin{equation}
\begin{aligned}
\mathcal{L}_{\mathrm{task}}
={}& \alpha\,\operatorname{CE}\bigl(y,p_S^{(1)}\bigr) \\
&+ (1-\alpha)\,T^2\,
\operatorname{KL}\bigl(p_C^{(T)}\,\|\,p_S^{(T)}\bigr),
\end{aligned}
\label{eq:goibert-task-loss}
\end{equation}
with $\alpha=0.5$ and $T=2$.
Because the supervision is at the output level, the tokenizer mismatch
between CafeBERT and GoiBERT needs no vocabulary alignment
\cite{boizard2025uld}.
For sequence classification and NLI, Eq.~\ref{eq:goibert-task-loss} is
applied once per example.
For ABSA, it is applied independently to each aspect slot.
For NER, both models' predictions are aligned to the first subword of
each original syllable, which yields a shared unit of supervision across both tokenizers.

\paragraph{Layer-wise learning-rate decay.}
During task fine-tuning, layer $\ell$ of the student uses learning rate
$\eta_\ell=\eta\,\gamma^{L_v-\ell}$ with $\gamma=0.8$, the embeddings
use $\eta\gamma^{L_v}$, and the task head uses $\eta$
\cite{howard2018universal,sun2019finetune}.
Lower layers, which hold the most transferable features learned in
Stage~1, are therefore updated more conservatively than upper,
task-specific ones.

\section{Experimental Setup}
\label{sec:setup}

\paragraph{Benchmarks.}
We evaluate on eight Vietnamese tasks that span four task families and
several domains (Appendix Table~\ref{tab:benchmarks}).
\emph{Text classification} covers sentiment and topic classification of
student feedback in UIT-VSFC \cite{nguyen2018uitvsfc} and four-class
spam detection of e-commerce reviews in ViSpamReviews
\cite{dinh2022vispam}.
\emph{Aspect-based sentiment analysis} (ABSA) covers smartphone reviews
in UIT-ViSFD \cite{phan2021sa2sl} and the Restaurant and Hotel domains of
UIT-ABSA \cite{dang2021uitabsa}.
\emph{Natural language inference} (NLI) uses ViNLI
\cite{huynh2022vinli} (the four-label community mirror
\texttt{presencesw/vinli\_4\_label}).
\emph{Named entity recognition} (NER) uses PhoNER\_COVID19
\cite{truong2021covid}.
We keep the official training, validation, and test splits.

\paragraph{Metrics.}
Classification and NLI are scored with accuracy and macro-F1, and ABSA
with aspect-detection F1 and end-to-end aspect--sentiment pair
micro-F1. NER is scored with strict entity-level F1.
This yields 15 metrics in total (Appendix~\ref{app:full-results}).
In the main text we report one primary metric per task: macro-F1 for
classification and NLI, pair F1 for ABSA, and entity F1 for NER.
All locally trained models are averaged over three fine-tuning seeds
(42, 43, 44) \cite{dodge2020finetuning}.

\paragraph{Baselines.}
We compare against the monolingual encoders PhoBERT-base-v2
\cite{nguyen2020phobert}, ViSoBERT \cite{nguyen2023visobert}, and
BamiBERT \cite{nguyen2026bamibert}; the multilingual XLM-R-base
\cite{conneau2020xlmr}; and CafeBERT \cite{do2024vlue}, which is also
our task teacher.
BamiBERT is our principal comparison because it is the most recent
compact Vietnamese encoder and our Stage~1 teacher.
We re-run it under the same downstream protocol as GoiBERT, with its
learning rate tuned on validation data and standard cross-entropy
fine-tuning.
ViDeBERTa \cite{tran2023videberta} results are reported in
Appendix~\ref{app:full-results}.

\paragraph{Distillation corpus.}
General distillation uses 328.6M tokens of clean, unlabeled Vietnamese
text from HPLT~3.0 \cite{oepen2026hplt}, Wikipedia, and the
\emph{training} splits of selected benchmarks.
To prevent leakage, we remove every passage that matches validation or
test content under exact matching or 13-gram overlap
\cite{brown2020fewshot,lee2022dedup}.

\paragraph{Implementation.}
General distillation runs for 120K steps (batch 32, sequence length
256) with AdamW, learning rate $10^{-4}$, and linear decay, on a single
NVIDIA L40S GPU in bfloat16.
Task fine-tuning runs for 30 epochs with batch size 32.
Learning rates and checkpoints are selected on validation data only.
Efficiency is measured on the encoder alone.
GPU throughput and peak memory are measured on an L40S (bf16, batch 32,
length 256), and CPU latency on a single 128-token input with eight
threads (fp32).
The same protocol is applied to every model.
Full hyperparameters are given in Appendix~\ref{app:training-details}.

\section{Results and Analysis}
\label{sec:results}

\begin{table*}[t]
\centering
\small
\begin{tabular*}{\textwidth}{@{\extracolsep{\fill}}llrrrrrrr@{}}
\toprule
\textbf{Task} & \textbf{Metric}
 & \textbf{Cafe-}
 & \textbf{XLM-R}$^{\dagger}$
 & \textbf{ViSo-}
 & \textbf{PhoBERT}
 & \textbf{Bami-}
 & \textbf{GoiBERT}
 & \textbf{GoiBERT} \\
 & & \textbf{BERT}$^{\diamond}$ & & \textbf{BERT} & \textbf{-v2} & \textbf{BERT} & \textbf{-base} & \textbf{-large} \\
\midrule
ViNLI      & M-F1    & 86.19 & 77.01 & 67.67 & 78.87 & \textbf{80.18} & 75.86 & 78.52 \\
ViSpam     & M-F1    & 77.22 & 76.55 & 77.10 & 77.86 & 77.55 & 76.52 & \textbf{78.13} \\
PhoNER     & Ent.\ F1& 95.47 & 92.50 & 93.07 & \textbf{95.44}$^{\ddagger}$ & 94.32 & 92.97 & 93.47 \\
ABSA-Rest. & Pair    & 77.41 & 71.58 & 74.68 & 74.12 & 73.53 & 74.15 & \textbf{76.00} \\
ABSA-Hotel & Pair    & 74.78 & 71.23 & 74.17 & 72.35 & 73.94 & 75.01 & \textbf{75.67} \\
VSFC-Topic & M-F1    & 79.81 & 79.56 & 80.00 & 79.68 & 78.88 & 80.17 & \textbf{80.42} \\
VSFC-Sent. & M-F1    & 82.91 & 82.20 & 80.69 & \textbf{83.34} & 82.36 & 81.50 & 82.03 \\
ViSFD      & Pair    & 85.98 & 65.81 & 83.77 & 84.05 & 84.41 & 83.94 & \textbf{84.78} \\
\midrule
\multicolumn{2}{l}{\textit{Average (8)}}
           & 82.47 & 77.06 & 78.89 & 80.71 & 80.65 & 80.02 & \textbf{81.13} \\
\midrule
\multicolumn{2}{l}{Params (M) $\downarrow$}
           & 558.9 & 277.5 & 97.0 & 134.4 & 102.4 & \textbf{26.4} & 49.4 \\
\multicolumn{2}{l}{GFLOPs (128 tok.) $\downarrow$}
           & 78.9 & 22.4 & 22.4 & 22.4 & 22.4 & \textbf{4.8} & 10.1 \\
\bottomrule
\end{tabular*}
\caption{Test results (\%) on eight Vietnamese benchmarks, one primary
metric per task, averaged over three seeds. M-F1: macro-F1; Pair:
aspect--sentiment pair F1; Ent.\ F1: strict entity-level F1.
\textbf{Bold} marks the best \emph{non-teacher} model in each row.
$^{\diamond}$Task teacher of GoiBERT (shown for reference).
$^{\dagger}$Reported results. $^{\ddagger}$Uses word-segmented input.
All 15 metrics and seed ranges appear in
Appendices~\ref{app:full-results}--\ref{app:seed-results}.}
\label{tab:main}
\end{table*}

\subsection{Main Results}
\label{sec:main-results}

Table~\ref{tab:main} compares GoiBERT with existing Vietnamese and
multilingual encoders.
We highlight three findings.

\paragraph{GoiBERT-large is the strongest non-teacher model.}
GoiBERT-large reaches an average of 81.13, ahead of PhoBERT-v2 (80.71)
and BamiBERT (80.65), and it is the best non-teacher model on five of
the eight tasks.
The largest gains over BamiBERT, its Stage-1 teacher, appear on
fine-grained sentiment: $+$2.47 pair F1 on ABSA-Restaurant and $+$1.73
on ABSA-Hotel, followed by $+$1.54 macro-F1 on VSFC-Topic.
Thus a student with 48\% of BamiBERT's parameters does not merely
approximate its teacher but outperforms it on most tasks.
The improvement comes from the second, CafeBERT-guided stage
(\S\ref{sec:ablation}).
The full 15-metric average gives the same ordering: 84.02 for
GoiBERT-large, 83.61 for BamiBERT, and 82.98 for GoiBERT-base
(Table~\ref{tab:full}).

\paragraph{GoiBERT-base is a strong extreme-efficiency point.}
With only 26.4M parameters and 4.8 GFLOPs, GoiBERT-base averages
80.02, which is 99.2\% of BamiBERT's average with a quarter of its size.
It still outperforms BamiBERT on both UIT-ABSA domains and on
VSFC-Topic, and it exceeds ViSoBERT and XLM-R-base on average despite
being 3.7$\times$ and 10.5$\times$ smaller, respectively.

\paragraph{The cost of compression is task-dependent.}
GoiBERT does not lead everywhere.
BamiBERT remains better on ViNLI ($-$1.66 for GoiBERT-large) and
PhoNER ($-$0.85).
PhoBERT-v2 is best among non-teacher models on PhoNER and
VSFC-Sentiment, although on PhoNER it benefits from external word
segmentation, which GoiBERT does not use.
We analyze these patterns next.

\subsection{Where Does Compression Hurt?}
\label{sec:task-analysis}

\begin{figure}[t]
  \centering
  \includegraphics[width=\columnwidth]{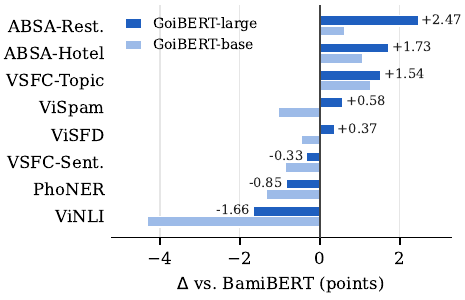}
  \caption{Per-task difference from BamiBERT (primary metric).
  Gains concentrate on ABSA and topic classification, whereas NLI and
  NER carry the cost of compression, most visibly for the smaller
  GoiBERT-base.}
  \label{fig:delta}
\end{figure}

\begin{table}[t]
\centering
\small
\setlength{\tabcolsep}{4pt}
\begin{tabular*}{\columnwidth}{@{\extracolsep{\fill}}lrrrr@{}}
\toprule
\textbf{Family} & \textbf{Bami} & \textbf{Goi-L} & \textbf{Cafe} & \textbf{$\Delta_{\text{L}-\text{C}}$} \\
\midrule
Classification (3) & 79.60 & 80.19 & 79.98 & $+$0.21 \\
ABSA (3)           & 77.29 & 78.82 & 79.39 & $-$0.57 \\
NER (1)            & 94.32 & 93.47 & 95.47 & $-$2.00 \\
NLI (1)            & 80.18 & 78.52 & 86.19 & $-$7.67 \\
\bottomrule
\end{tabular*}
\caption{Primary scores averaged by task family. $\Delta_{\text{L}-\text{C}}$
is the gap between GoiBERT-large and its 11$\times$ larger task teacher
CafeBERT.}
\label{tab:family}
\end{table}

Figure~\ref{fig:delta} shows the per-task differences from BamiBERT,
and Table~\ref{tab:family} groups tasks by family and compares
GoiBERT-large with both of its teachers.
The task ranking is nearly identical for both variants, so the pattern
reflects the tasks rather than single-model noise.
A clear pattern emerges.
On \emph{classification}, GoiBERT-large slightly \emph{surpasses} its
559M-parameter task teacher on average (80.19 vs.\ 79.98).
It also exceeds the teacher on three individual tasks (ViSpam,
ABSA-Hotel, and VSFC-Topic).
On \emph{ABSA}, it stays within 0.6 points of the teacher.
This is consistent with prior observations that soft targets act as a
data-dependent regularizer and can allow students to match or surpass
their teachers \cite{furlanello2018born,yuan2020revisiting}.
Out-of-fold targets may strengthen this effect, since they encode the
teacher's uncertainty on unseen examples rather than its memorized
labels.
Excluding ViNLI, GoiBERT-large averages 81.50 over the remaining seven
tasks, only 0.44 points below CafeBERT (81.94), while being
11$\times$ smaller in parameter count.

The remaining gap lies almost entirely in \emph{NLI} and, to a lesser
extent, \emph{NER}.
Two observations suggest that the ViNLI gap reflects limited capacity
rather than a lack of supervision.
First, ViNLI is the task most sensitive to model size within the GoiBERT
family: the base--large gap is 2.66 points, compared with 0.89 on
average across the other tasks.
Second, ViNLI gains $+$1.4 (large) and $+$3.0 (base) points from task
distillation (\S\ref{sec:ablation}), so the teacher's signal is
informative, but a 12-layer, 512-dimensional
student cannot fully absorb the cross-sentence reasoning that the
24-layer, 1024-dimensional teacher performs.
For NER, two factors may contribute.
Entity F1 is already high (above 92 for all models), leaving little
room for distillation to help.
In addition, aligning supervision to the first subword of each syllable
discards the multi-syllable word boundaries that PhoBERT receives from
its segmenter, and these boundaries frequently coincide with entity
boundaries in Vietnamese.
Distillation objectives tailored to inference and span structure are a
natural next step.

\subsection{Efficiency--Accuracy Trade-off}
\label{sec:efficiency-analysis}

\begin{table}[t]
\centering
\footnotesize
\setlength{\tabcolsep}{2pt}
\begin{tabular*}{\columnwidth}{@{\extracolsep{\fill}}lrrrr@{}}
\toprule
\textbf{Model} & \textbf{Weights} & \textbf{Throughput} & \textbf{VRAM} & \textbf{CPU lat.} \\
\midrule
BamiBERT  & 206 MB & 2,397 \phantom{\scriptsize(0.0$\times$)} & 801 MB & 31.0 ms \\
Goi-large & 99 MB  & 4,827 {\scriptsize(2.0$\times$)} & 410 MB & 19.4 ms \\
Goi-base  & \textbf{53 MB} & \textbf{8,285} {\scriptsize(3.5$\times$)} & \textbf{248 MB} & \textbf{14.2 ms} \\
\bottomrule
\end{tabular*}
\caption{Inference efficiency of the encoders (bf16 weights).
Throughput (sentences/s) and peak VRAM: L40S, batch 32, length 256.
CPU latency: one 128-token input, eight threads, fp32.
Throughput speed-ups relative to BamiBERT are in parentheses.}
\label{tab:efficiency}
\end{table}

Table~\ref{tab:efficiency} turns the parameter and FLOP reductions of
Table~\ref{tab:main} into measured deployment gains.
Compared with BamiBERT, GoiBERT-large doubles GPU throughput, halves
peak memory, and cuts CPU latency by 37\%.
GoiBERT-base processes 3.5$\times$ as many sentences per second and
needs less than a third of the memory.
Together with Figure~\ref{fig:teaser}, these results place both
variants on the accuracy--efficiency frontier.
GoiBERT-large dominates every baseline except the 11$\times$ larger
CafeBERT, and GoiBERT-base is the only model below 30M parameters.

The measured speed-ups are, however, smaller than the FLOP ratios
suggest.
GoiBERT-base needs 4.7$\times$ fewer FLOPs than BamiBERT but runs
only 3.5$\times$ faster on GPU and 2.2$\times$ faster on CPU.
At small widths, memory-bound operations such as layer normalization
and embedding lookups, together with fixed per-call overhead, take up a
growing share of runtime.
This echoes the warning that FLOPs and parameter counts are imperfect
proxies for real efficiency \cite{dehghani2022efficiency}.
It also makes GoiBERT-large, whose 2.2$\times$ FLOP reduction
translates almost fully into a 2.0$\times$ GPU speed-up, the more
balanced choice when accuracy matters.
GoiBERT-base is preferable for CPU-only or memory-constrained serving.

\paragraph{Where do the parameters go?}
Table~\ref{tab:full} breaks each model's size into embedding and
encoder parameters, which explains much of the efficiency picture.
Multilingual vocabularies impose a large fixed cost:
69\% of XLM-R-base and 46\% of CafeBERT are embeddings, and these
parameters add memory without adding per-token computation.
BamiBERT's compact 20K-token vocabulary already avoids most of this
cost, so GoiBERT's savings over BamiBERT come mainly from the encoder:
37.8M encoder parameters for GoiBERT-large and 17.7M for GoiBERT-base,
compared with 85.1M.
Consequently, the encoder is where further compression should target,
for example through pruning or quantization
\cite{xia2022cofi,kim2021ibert}, whereas for multilingual models
vocabulary trimming is the more natural first step
\cite{abdaoui2020loadwhat,ushio2023vocabtrim}.

\paragraph{Which variant should practitioners use?}
The two variants serve different deployment regimes.
When accuracy is the priority and a GPU is available, GoiBERT-large is
the better default.
It is the strongest non-teacher model on average, roughly matches
CafeBERT on classification and ABSA, and still halves BamiBERT's
serving cost.
When the budget is tight, such as CPU-only inference, edge devices, or
high-throughput batch labeling, GoiBERT-base offers the lowest latency
and memory with an average loss of only 1.1 points relative to
GoiBERT-large.
For NLI-heavy applications, a larger encoder remains preferable.

\subsection{Ablation Study}
\label{sec:ablation}

\begin{table}[t]
\centering
\small
\setlength{\tabcolsep}{3.5pt}
\begin{tabular*}{\columnwidth}{@{\extracolsep{\fill}}lccc@{}}
\toprule
\textbf{Setting} & \textbf{ABSA-R} & \textbf{VSFC-T} & \textbf{ViNLI} \\
\midrule
\multicolumn{4}{l}{\textit{GoiBERT-large}} \\
\quad CE only (no GD)   & 59.35 & 77.42 & 45.02 \\
\quad + General KD      & 72.89 {\scriptsize$+$13.5} & 78.06 {\scriptsize$+$0.6} & 76.18 {\scriptsize$+$31.2} \\
\quad + Task KD (OOF)   & 74.81 {\scriptsize$+$1.9} & 78.94 {\scriptsize$+$0.9} & 77.60 {\scriptsize$+$1.4} \\
\quad + LLRD (full)     & \textbf{76.00} {\scriptsize$+$1.2} & \textbf{80.42} {\scriptsize$+$1.5} & \textbf{78.52} {\scriptsize$+$0.9} \\
\midrule
\multicolumn{4}{l}{\textit{GoiBERT-base}} \\
\quad CE only (no GD)   & 59.43 & 77.30 & 44.04 \\
\quad + General KD      & 71.12 {\scriptsize$+$11.7} & 79.29 {\scriptsize$+$2.0} & 71.10 {\scriptsize$+$27.1} \\
\quad + Task KD (OOF)   & 72.85 {\scriptsize$+$1.7} & 78.72 {\scriptsize$-$0.6} & 74.05 {\scriptsize$+$3.0} \\
\quad + LLRD (full)     & \textbf{74.15} {\scriptsize$+$1.3} & \textbf{80.17} {\scriptsize$+$1.5} & \textbf{75.86} {\scriptsize$+$1.8} \\
\bottomrule
\end{tabular*}
\caption{Cumulative ablation (test, mean of three seeds).
ABSA-R: Restaurant pair F1; VSFC-T and ViNLI: macro-F1.
Small numbers give the change from the previous row.}
\label{tab:ablation}
\end{table}

Table~\ref{tab:ablation} adds each component in turn on three tasks
that span ABSA, topic classification, and NLI.

\paragraph{General distillation is indispensable.}
Training the compact architecture from scratch fails on tasks that
require compositional understanding: ViNLI macro-F1 is only 45.02 and
ABSA-Restaurant pair F1 is 59.35.
Stage~1 recovers $+$31.2 and $+$13.5 points on these tasks,
respectively, for GoiBERT-large.
In contrast, VSFC-Topic gains less than one point, since lexical cues
already go a long way for topic classification.
The magnitude of these gains confirms that representation transfer from
a tokenizer-compatible teacher is the backbone of the recipe
\cite{turc2019wellread,jiao2020tinybert}.

\paragraph{Cross-tokenizer task distillation adds consistent gains.}
Out-of-fold CafeBERT supervision improves five of the six settings, by
up to $+$3.0 points (ViNLI, GoiBERT-base).
For GoiBERT-base, its largest effect is on ViNLI, the task with the
widest teacher--student gap.
This supports our design choice to bring in the stronger teacher at
the stage where its tokenizer does not matter.
The one decrease ($-$0.6 on VSFC-Topic for GoiBERT-base) is small and
is more than recovered once LLRD is subsequently added.

\paragraph{LLRD helps uniformly.}
Layer-wise decay improves every task for both variants, by 0.9--1.8
points.
This is consistent with the intuition that the lower layers distilled
in Stage~1 should be preserved during fine-tuning
\cite{howard2018universal,zhang2021revisiting}.
Overall, the full pipeline improves on cross-entropy fine-tuning of an
undistilled student by 16.7, 3.0, and 33.5 points for GoiBERT-large.

\paragraph{Stability across seeds.}
GoiBERT-large is also more stable than BamiBERT under fine-tuning.
Its seed range (max$-$min) is narrower on 11 of the 15 metrics, and
its average range is 0.61 points versus 0.85 for BamiBERT
(Appendix~\ref{app:seed-results}).
This is in line with the regularizing effect of soft targets
\cite{muller2019labelsmoothing}.

\section{Conclusion}
\label{sec:conclusion}

We introduced GoiBERT, a family of compact, segmentation-free
Vietnamese encoders trained by two-stage distillation from two
complementary teachers.
A tokenizer-compatible teacher transfers general representations, and a
much larger cross-tokenizer teacher supplies out-of-fold task
supervision.
GoiBERT-large (49.4M parameters) achieves the best average score among
non-teacher models across eight benchmarks, doubles the GPU throughput
of BamiBERT, and comes within half a point of its 559M-parameter teacher
outside NLI.
GoiBERT-base (26.4M) retains 99\% of BamiBERT's average with a quarter
of its parameters.
Our analysis shows that compression is nearly free for classification
and ABSA but still costly for inference and entity recognition.
This two-teacher recipe should also transfer to other languages whose
strongest models are multilingual.
Future work will target NLI- and NER-specific distillation, longer
contexts, unseen domains, and quantization.

\section*{Limitations}

Our comparison with BamiBERT is not a pure architecture comparison.
GoiBERT also benefits from CafeBERT distillation, layer-wise
learning-rate decay, and a general-distillation corpus that includes
benchmark \emph{training} text, while our BamiBERT re-run uses plain
cross-entropy fine-tuning. We should also be upfront that we decided to build the larger variant after looking at earlier test results, so test feedback partly shaped how the models were developed.

On the statistical side, we ran general distillation only once.
The seed ranges we report therefore reflect fine-tuning variability
alone, and they are not confidence intervals or significance tests.
Our ablations cover just three tasks and add components one at a time, so they do not fully separate how the components interact.
The 8- and 15-metric averages are handy summaries, but they mix metrics that mean different things.

There are also things we did not test. We distill on sequences of at most 256 tokens, so we do not know how GoiBERT handles long inputs, and none of our benchmarks probe unseen domains or dialectal variation. Our efficiency numbers cover encoder inference on the hardware we used. They leave out tokenization, task heads, and the cost of training five out-of-fold CafeBERT teachers for each task, which is substantial even though it is paid only once.

\section*{Ethics Statement}
GoiBERT is trained on publicly available text and benchmarks, so it can pick up social biases from web data and from its teachers.
We have not audited the models for such biases, and we encourage
anyone deploying them in sensitive settings to check first.
On the positive side, smaller models make Vietnamese NLP cheaper to run and accessible on more modest hardware.

\bibliography{custom}

\appendix
\null
\newpage
\section{Architecture and Initialization}
\label{app:architecture}

\begin{figure}[!t]
\centering
\includegraphics[width=0.8\columnwidth]
{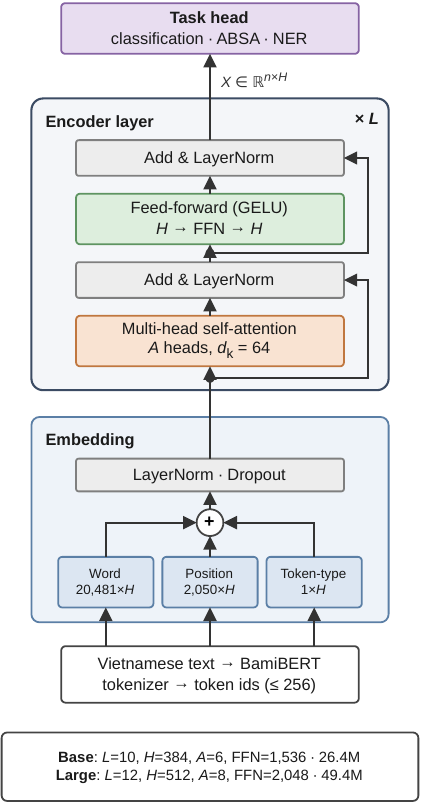}
\caption{
Detailed encoder architecture of GoiBERT-base and GoiBERT-large.
}
\label{fig:goibert_architecture}
\end{figure}

Figure~\ref{fig:goibert_architecture} provides a layer-level view of
the GoiBERT encoder family.
The model configurations are summarized in \S\ref{sec:arch}; here we
provide the additional implementation details omitted from the main paper.

For GoiBERT-large, the word embeddings are initialized from BamiBERT's
768-dimensional embedding matrix using principal component analysis
(PCA), retaining the leading 512 components.
The projected embeddings are normalized and rescaled to match the
standard deviation of the student's embedding initialization.
Transformer blocks are initialized randomly.
Both variants use BamiBERT's tokenizer and support up to 2,050 position
embeddings; general distillation uses sequences of at most 256 tokens.

\section{Distillation Details}
\label{app:distillation}

This section expands the matching terms used in the general-distillation
objective in Eq.~\ref{eq:goibert-general-loss} (\S\ref{sec:gd}) and the optimization
schedule used during task-specific distillation.
Let $v\in\{\mathrm{base},\mathrm{large}\}$ index a GoiBERT variant,
and let $L_v$, $d_v$, and $a_v$ denote its number of Transformer layers,
hidden dimension, and number of attention heads, respectively.

\paragraph{Hidden-state matching.}
For a sequence of $n$ tokens, let
$H_\ell^v\in\mathbb{R}^{n\times d_v}$ denote the student hidden states
at layer $\ell$, with $\ell=0$ denoting the embedding output.
Let $H_{\pi_v(\ell)}^B\in\mathbb{R}^{n\times768}$ denote the hidden
states of the aligned BamiBERT teacher layer.
The function $\pi_v(\ell)$ specifies which teacher layer is paired with
student layer $\ell$.

\paragraph{Layer mapping.}
Hidden-state distillation needs one teacher target for every student
layer.
For GoiBERT-large, both encoders have 12 Transformer layers, so the
mapping is one-to-one: $\pi_{\mathrm{large}}(\ell)=\ell$ for
$\ell=0,\ldots,12$.
GoiBERT-base has 10 layers and cannot be paired one-to-one with the 12
teacher layers, so we use a layer mapping
$\pi_{\mathrm{base}}:\{0,\ldots,10\}\rightarrow\{0,\ldots,12\}$ that
selects 10 of the teacher's hidden states as targets.
The mapping fixes $\pi_{\mathrm{base}}(0)=0$, so that student and teacher
embedding outputs are aligned, and it is strictly increasing, so that
the depth order of the teacher is preserved: each student layer
imitates a teacher layer at a comparable stage of processing, and the
two unselected teacher layers receive no matching term.
This follows the layer-mapping strategy of TinyBERT
\cite{jiao2020tinybert}, where a shallower student learns from a
subset of teacher layers.
In our implementation, the selected teacher layers are
$\bigl(\pi_{\mathrm{base}}(1),\ldots,\pi_{\mathrm{base}}(10)\bigr)
=(1,2,4,5,6,7,8,10,11,12)$, together with $\pi_{\mathrm{base}}(0)=0$
for the embedding outputs.
GoiBERT-base therefore skips teacher layers 3 and 9, while the
remaining layers keep the teacher's depth order.
Each student hidden state is mapped from 384 to 768 dimensions by its
own learned linear projection, and the attention-matching term below
uses the same Transformer-layer alignment.

Because the hidden dimensions differ, a trainable linear projection
$P_\ell^v:\mathbb{R}^{d_v}\rightarrow\mathbb{R}^{768}$ maps the
student representation to the teacher space before comparison.
The hidden-state term is

\begin{equation}
  \mathcal{L}_{\mathrm{hid}}
  = \frac{1}{L_v+1}\sum_{\ell=0}^{L_v}
  \operatorname{MSE}_{\mathrm{nonpad}}\!\bigl(P^v_\ell(H^v_\ell),\,H^B_{\pi_v(\ell)}\bigr),
\end{equation}

Here, $\operatorname{MSE}_{\mathrm{nonpad}}$ denotes mean-squared error evaluated only at non-padding token positions. Padding tokens therefore do not contribute to the loss. The projection layers are optimized jointly with the student encoder, while BamiBERT remains frozen.

\paragraph{Attention matching.}
The student and teacher also use different numbers of attention heads.
Direct head-to-head matching would therefore be undefined.
Instead, we average the post-softmax attention probabilities within
each layer before comparing the two models.

Let
$A_{\ell,h}^{m}\in\mathbb{R}^{n\times n}$
be the attention probability matrix produced by head $h$ of layer
$\ell$ in model $m$, and let $a_m$ be the number of attention heads
in that model.
We define the head-averaged attention matrix as

\begin{equation}
\bar{A}_{\ell}^{m}
=
\frac{1}{a_m}
\sum_{h=1}^{a_m}
A_{\ell,h}^{m}.
\nonumber
\end{equation}

Each entry of $\bar{A}_{\ell}^{m}$ therefore represents the average
attention probability assigned from one token position to another
across all heads in that layer.
The attention-matching term is

\begin{equation}
\mathcal{L}_{\mathrm{att}}^{v}
=\dfrac{1}{L_v}
\sum_{\ell=1}^{L_v}
\operatorname{MSE}
\left(
\bar{A}_{\ell}^{v},
\bar{A}_{\pi_v(\ell)}^{B}
\right).
\nonumber
\end{equation}

The summation starts at the first Transformer layer because the
embedding output has no self-attention matrix.
The hidden-state and attention terms in
Eqs.~\ref{eq:hid}--\ref{eq:att} are combined as
specified by Eq.~\ref{eq:goibert-general-loss} in the main paper.

\paragraph{Out-of-fold soft supervision.}
The task-specific objective itself is defined in
Eq.~\ref{eq:goibert-task-loss}.
Here we clarify how its CafeBERT soft targets are produced.
The task training set is divided into five folds.
For fold $k$, a CafeBERT teacher is fine-tuned on the other four folds
and used only to predict examples in the held-out fold.
Repeating this procedure over all five folds gives one soft target for
every training example without allowing its teacher to train on that
example.

The resulting CafeBERT logits are converted to temperature-scaled
probabilities as defined in Eq.~\ref{eq:goibert-task-loss}, using
$T=2$ and $\alpha=0.5$.
For ABSA, distillation is applied independently to each aspect slot.
For NER, CafeBERT soft targets are aligned with the first subword of
each original syllable.

\paragraph{Layer-wise learning-rate decay.}
During task-specific fine-tuning, lower encoder layers receive
progressively smaller updates.
Let $\eta$ denote the task-level base learning rate and let
$\ell=1,\ldots,L_v$ index Transformer layers from bottom to top.
The learning rate assigned to layer $\ell$ is

\begin{equation}
\eta_\ell
=
\eta\,\gamma^{L_v-\ell},
\qquad
\gamma=0.8.
\label{eq:app-llrd}
\end{equation}

Thus, the top Transformer layer receives $\eta$, while layers closer
to the embedding layer receive increasingly smaller rates.
The embedding parameters use $\eta\gamma^{L_v}$, and the task-specific
head uses the full base rate $\eta$.

\begin{table*}[!t]
\centering
\footnotesize

\begin{tabular*}{\textwidth}
{@{\extracolsep{\fill}}llrrrrrrrr@{}}
\toprule
\textbf{Task} & \textbf{Metric}
 & \textbf{Cafe-}
 & \textbf{XLM-R}$^{\dagger}$
 & \textbf{ViDe-}
 & \textbf{ViSo-}
 & \textbf{PhoBERT}
 & \textbf{Bami-}
 & \textbf{GoiBERT}
 & \textbf{GoiBERT} \\
 & & \textbf{BERT}
 &
 & \textbf{BERTa}$^{\dagger}$
 & \textbf{BERT}
 & \textbf{-v2}
 & \textbf{BERT}
 & \textbf{-base}
 & \textbf{-large} \\
\midrule

ViNLI
& Acc.
& \textbf{86.08}
& 76.83
& 61.08
& 67.55
& 78.75
& 80.07
& 75.75
& 78.39 \\

&
M-F1
& \textbf{86.19}
& 77.01
& 60.71
& 67.67
& 78.87
& 80.18
& 75.86
& 78.52 \\

\addlinespace[1pt]

ViSpam
& Acc.
& 90.50
& 90.16
& 86.21
& 90.78
& \textbf{90.82}
& 90.19
& 89.88
& 90.75 \\

&
M-F1
& 77.22
& 76.55
& 67.04
& 77.10
& 77.86
& 77.55
& 76.52
& \textbf{78.13} \\

\addlinespace[1pt]

PhoNER
& Ent.\ F1
& \textbf{95.47}
& 92.50
& 94.50$^{\ddagger}$
& 93.07
& 95.44$^{\ddagger}$
& 94.32
& 92.97
& 93.47 \\

\addlinespace[1pt]

ABSA-Rest.
& Det.
& \textbf{88.83}
& 82.18
& 73.56
& 87.06
& 87.32
& 87.03
& 86.91
& 88.14 \\

&
Pair
& \textbf{77.41}
& 71.58
& 63.78
& 74.68
& 74.12
& 73.53
& 74.15
& 76.00 \\

\addlinespace[1pt]

ABSA-Hotel
& Det.
& 80.37
& 77.70
& 72.05
& 79.94
& 78.94
& 80.15
& 80.66
& \textbf{80.96} \\

&
Pair
& 74.78
& 71.23
& 62.97
& 74.17
& 72.35
& 73.94
& 75.01
& \textbf{75.67} \\

\addlinespace[1pt]

VSFC-Topic
& Acc.
& 88.88
& 89.18
& 83.94
& 88.85
& 88.89
& 88.82
& 89.36
& \textbf{89.73} \\

&
M-F1
& 79.81
& 79.56
& 66.49
& 80.00
& 79.68
& 78.88
& 80.17
& \textbf{80.42} \\

\addlinespace[1pt]

VSFC-Sent.
& Acc.
& \textbf{93.79}
& 93.56
& 87.86
& 92.57
& 93.63
& 93.50
& 92.70
& 93.56 \\

&
M-F1
& 82.91
& 82.20
& 70.74
& 80.69
& \textbf{83.34}
& 82.36
& 81.50
& 82.03 \\

\addlinespace[1pt]

ViSFD
& Det.
& \textbf{90.05}
& 82.73
& 75.53
& 88.93
& 89.32
& 89.26
& 89.30
& 89.81 \\

&
Pair
& \textbf{85.98}
& 65.81
& 64.20
& 83.77
& 84.05
& 84.41
& 83.94
& 84.78 \\

\midrule

\multicolumn{2}{l}{\textbf{Average (15)}}
& \textbf{85.22}
& 80.59
& 72.71
& 81.79
& 83.56
& 83.61
& 82.98
& 84.02 \\

\midrule

\multicolumn{2}{l}{Params (M)}
& 558.9
& 277.5
& $\sim$183.4
& 97.0
& 134.4
& 102.4
& \textbf{26.4}
& 49.4 \\

\multicolumn{2}{l}{\quad embeddings}
& 256.5
& 192.4
& $\sim$98.3
& 11.9
& 49.4
& 17.3
& \textbf{8.7}
& 11.5 \\

\multicolumn{2}{l}{\quad encoder}
& 302.3
& 85.1
& $\sim$85.1
& 85.1
& 85.1
& 85.1
& \textbf{17.7}
& 37.8 \\

\midrule

\multicolumn{2}{l}{GFLOPs (128 tok.)}
& 78.9
& 22.4
& $\geq$22.4
& 22.4
& 22.4
& 22.4
& \textbf{4.8}
& 10.1 \\

\multicolumn{2}{l}{GFLOPs (512 tok.)}
& 335.0
& 96.6
& $\geq$96.6
& 96.6
& 96.6
& 96.6
& \textbf{22.1}
& 45.1 \\

\bottomrule
\end{tabular*}

\caption{
Complete test results (\%) across all 15 evaluation metrics.
Bold indicates the best value in each row; lower is better for
parameter counts and GFLOPs.
$^{\dagger}$ denotes reported results and
$^{\ddagger}$ denotes results requiring word-segmented input.
Approximate values and lower bounds are marked by $\sim$ and $\geq$,
respectively.
ViNLI uses the community mirror
\texttt{presencesw/vinli\_4\_label}.
}
\label{tab:full}
\end{table*}

\begin{table}[!t]
\centering
\footnotesize
\setlength{\tabcolsep}{2.2pt}
\renewcommand{\arraystretch}{1.02}

\begin{tabular*}{\columnwidth}
{@{\extracolsep{\fill}}llcc@{}}
\toprule
\textbf{Task}
& \textbf{Metric}
& \textbf{BamiBERT}
& \textbf{GoiBERT-large} \\
\midrule

ViNLI
& Acc.
& 80.07 {\scriptsize[79.94, 80.24]}
& 78.39 {\scriptsize[78.33, 78.47]} \\

&
M-F1
& 80.18 {\scriptsize[80.04, 80.33]}
& 78.52 {\scriptsize[78.46, 78.57]} \\

\addlinespace[0.5pt]

ViSpam
& Acc.
& 90.19 {\scriptsize[89.98, 90.41]}
& 90.75 {\scriptsize[90.44, 91.07]} \\

&
M-F1
& 77.55 {\scriptsize[75.58, 78.62]}
& 78.13 {\scriptsize[77.78, 78.32]} \\

\addlinespace[0.5pt]

PhoNER
& Ent.\ F1
& 94.32 {\scriptsize[94.24, 94.39]}
& 93.47 {\scriptsize[93.44, 93.48]} \\

\addlinespace[0.5pt]

ABSA-Rest.
& Det.
& 87.03 {\scriptsize[86.45, 87.52]}
& 88.14 {\scriptsize[87.90, 88.39]} \\

&
Pair
& 73.53 {\scriptsize[72.36, 74.15]}
& 76.00 {\scriptsize[75.72, 76.52]} \\

\addlinespace[0.5pt]

ABSA-Hotel
& Det.
& 80.15 {\scriptsize[80.12, 80.19]}
& 80.96 {\scriptsize[80.72, 81.23]} \\

&
Pair
& 73.94 {\scriptsize[73.82, 74.06]}
& 75.67 {\scriptsize[75.27, 75.99]} \\

\addlinespace[0.5pt]

VSFC-Topic
& Acc.
& 88.82 {\scriptsize[88.50, 89.36]}
& 89.73 {\scriptsize[89.45, 89.96]} \\

&
M-F1
& 78.88 {\scriptsize[78.18, 79.80]}
& 80.42 {\scriptsize[79.88, 81.03]} \\

\addlinespace[0.5pt]

VSFC-Sent.
& Acc.
& 93.50 {\scriptsize[93.24, 93.75]}
& 93.56 {\scriptsize[93.40, 93.71]} \\

&
M-F1
& 82.36 {\scriptsize[81.61, 83.03]}
& 82.03 {\scriptsize[80.77, 83.36]} \\

\addlinespace[0.5pt]

ViSFD
& Det.
& 89.26 {\scriptsize[89.03, 89.39]}
& 89.81 {\scriptsize[89.64, 89.90]} \\

&
Pair
& 84.41 {\scriptsize[84.11, 84.64]}
& 84.78 {\scriptsize[84.66, 85.03]} \\

\bottomrule
\end{tabular*}

\caption{
Test-score variation for BamiBERT and GoiBERT-large.
Values are means with [min, max] across seeds 42, 43, and 44.
}
\label{tab:seeds}
\end{table}

\section{Benchmark Details}
\label{app:benchmarks}

Table~\ref{tab:benchmarks} summarizes the eight evaluation tasks.
We use the official training, validation, and test partitions of each
dataset version.

\begin{table}[ht]
\centering
\footnotesize
\setlength{\tabcolsep}{3pt}
\begin{tabular*}{\columnwidth}{@{\extracolsep{\fill}}llll@{}}
\toprule
\textbf{Task} & \textbf{Family} & \textbf{Domain} & \textbf{Metrics} \\
\midrule
VSFC-Sent.  & Classif. & Student feedback & Acc., M-F1 \\
VSFC-Topic  & Classif. & Student feedback & Acc., M-F1 \\
ViSpam      & Classif. & E-commerce reviews & Acc., M-F1 \\
ViSFD       & ABSA & Smartphone reviews & Det., Pair \\
ABSA-Rest.  & ABSA & Restaurant reviews & Det., Pair \\
ABSA-Hotel  & ABSA & Hotel reviews & Det., Pair \\
ViNLI       & NLI  & News (open domain) & Acc., M-F1 \\
PhoNER      & NER  & COVID-19 news & Ent.\ F1 \\
\bottomrule
\end{tabular*}
\caption{Evaluation benchmarks. Det.: aspect-detection F1; Pair:
aspect--sentiment pair micro-F1.}
\label{tab:benchmarks}
\end{table}

\section{Training and Measurement Details}
\label{app:training-details}

General distillation is run for 120,000 optimization steps with batch
size 32 and sequence length 256.
We use AdamW with learning rate $10^{-4}$, weight decay $0.01$,
1,000 warmup steps, and linear learning-rate decay.
This stage uses seed 42 and bfloat16 precision on one NVIDIA L40S GPU.

For downstream tasks, both GoiBERT variants are fine-tuned for
30 epochs with batch size 32 using AdamW, weight decay $0.01$, and
10\% warmup.
The base learning rate is $10^{-5}$ for the two UIT-VSFC tasks,
$5\times10^{-5}$ for ViSpamReviews, and $10^{-4}$ for the remaining
tasks.
GoiBERT additionally uses the layer-wise decay in
Eq.~\ref{eq:app-llrd}.
For the rerun BamiBERT baseline, the learning rate is selected from
$\{10^{-5},2\times10^{-5},5\times10^{-5}\}$ using validation
performance; BamiBERT is trained with cross-entropy only.

Model selection is based exclusively on validation data.
Reported local test results are averaged over fine-tuning seeds
42, 43, and 44.

Efficiency measurements use the encoder only.
GPU throughput and peak VRAM are measured on an NVIDIA L40S using
bfloat16 and scaled-dot-product attention at batch size 32 and sequence
length 256.
CPU latency is measured in float32 with eight CPU threads for a single
128-token input.
Timing measurements use repeated forward passes after warm-up; the
same measurement protocol is applied to all compared encoders.


\section{Complete Benchmark Results}
\label{app:full-results}

Table~\ref{tab:full} provides the complete set of 15 test metrics,
together with parameter counts and estimated computational cost.
The 15-metric average is reported only as a descriptive summary:
the metrics have different interpretations, and tasks contributing two
metrics receive greater weight than PhoNER, which contributes one.
The GoiBERT--BamiBERT comparison also reflects the complete downstream
training pipelines rather than architecture alone, since GoiBERT uses
CafeBERT distillation and layer-wise learning-rate decay whereas the
rerun BamiBERT baseline uses standard cross-entropy fine-tuning.

\section{Seed-wise Variation}
\label{app:seed-results}

Table~\ref{tab:seeds} reports variation across seeds 42, 43, and 44
for BamiBERT and GoiBERT-large.
The ranges favor GoiBERT-large on nine metrics, BamiBERT on both
ViNLI metrics and PhoNER entity $F_1$, and overlap on ViSpam
macro-$F_1$ and both VSFC-Sentiment metrics.
They describe fine-tuning variability only, not confidence intervals
or significance tests, and exclude repeated general distillation.

\end{document}